\documentclass[lettersize,journal]{IEEEtran}
\usepackage{amsmath,amsfonts}
\usepackage{array}
\usepackage[caption=false,font=normalsize,labelfont=sf,textfont=sf]{subfig}
\usepackage{textcomp}
\usepackage{stfloats}
\usepackage{url}
\usepackage{verbatim}
\usepackage{graphicx}
\usepackage{enumitem}

\usepackage{booktabs} 
\usepackage{multirow} 
\usepackage{titlesec}

\usepackage[linesnumbered, ruled,vlined]{algorithm2e}
\usepackage[colorlinks, linkcolor=blue, citecolor=blue, urlcolor=black]{hyperref}
\usepackage[table]{xcolor}

\def\BibTeX{{\rm B\kern-.05em{\sc i\kern-.025em b}\kern-.08em
    T\kern-.1667em\lower.7ex\hbox{E}\kern-.125emX}}
\usepackage{balance}
\begin{document}
\title{Visual Neural Decoding via Improved Visual-EEG Semantic Consistency}
\author{
Hongzhou~Chen, 
Lianghua~He,~\IEEEmembership{Member,~IEEE,}
Yihang~Liu, 
Longzhen~Yang,
Shaohua~Shang,
and MengChu~Zhou,~\IEEEmembership{Fellow,~IEEE}
\thanks{Hongzhou Chen, Yihang Liu, Longzhen Yang, and Shaohua Shang are with the School of Computer Science and Technique, Tongji University, Shanghai 201804, China (e-mail: chenhongzhou@tongji.edu.cn; 2111131@tongji.edu.cn; yanglongzhen@tongji.edu.cn; shaohuashang@tongji.edu.cn).}
\thanks{Lianghua He is with the Shanghai Eye Disease Prevention and Treatment Center, Shanghai 200040, China, and also with the School of Computer Science and Technique, Tongji University, Shanghai 201804, China (e-mail: Helianghua@tongji.edu.cn).}
\thanks{MengChu Zhou is with the Helen and John C. Hartmann Department of Electrical and Computer Engineering, New Jersey Institute of Technology, Newark, NJ 07102 USA (e-mail: zhou@njit.edu). }
\thanks{Lianghua He and MengChu Zhou are the corresponding authors.}

}


\maketitle

\begin{abstract}
Visual neural decoding aims to extract and interpret original visual experiences directly from human brain activity.
Recent studies have demonstrated the feasibility of decoding visual semantic categories from electroencephalography (EEG) signals, among which metric learning-based approaches have delivered promising results.
However, these methods that directly map EEG features into a pre-trained embedding space inevitably introduce mapping bias, resulting in a modality gap and semantic inconsistency that impair cross-modal alignment.
To address these issues, this work constructs a Visual-EEG Joint Semantic Space to bridge the gap between visual images and neural signals.
Building upon this space, we propose two novel approaches to improve semantic consistency between cross-modal representations and facilitate optimal alignment.
Specifically, (1) we introduce a Visual-EEG Semantic Decoupling Network (VE-SDN) to explicitly disentangle semantic components from modality representations, thereby achieving purely semantic-level cross-modal alignment.
(2) We introduce a Neural-Guided Intra-Class Consistency (NGIC) objective, an asymmetric representation alignment strategy designed to effectively enhance the robustness of visual representations and further boost decoding performance.
Extensive experiments on a large-scale Visual-EEG dataset validate the effectiveness of the proposed method. 
Compared to the strongest baseline, our approach demonstrates superior decoding performance, yielding relative Top-1/Top-5 accuracy improvements of 38.9\%/17.9\% in intra-subject and 16.1\%/11.3\% in inter-subject settings.
The code is available at \href{https://github.com/hzalanchen/Cross-Modal-EEG}
{https://github.com/hzalanchen/Cross-Modal-EEG}.
\end{abstract}


\section{Introduction}
\IEEEPARstart{A}{ccurate} decoding of visual information from brain activity remains a fundamental challenge in brain-computer interfaces (BCIs) \cite{zhang2022neural}.
Currently, much research is dedicated to decoding visual neural activity using functional magnetic resonance imaging (fMRI) \cite{chen2023seeing}, \cite{scotti2024mindeye2}, \cite{xia2024umbrae}; however, the exorbitant cost of equipment and data acquisition severely hinder its widespread adoption in real-world applications. 
In contrast, electroencephalography (EEG) rapidly emerges as a highly viable alternative, owing to its portability, cost-effectiveness, and superior temporal resolution that enables the precise capture of transient neural dynamics \cite{niedermeyer2005electroencephalography}, \cite{yen2023exploring}.
Recent studies confirm that extracting visual semantic categories from EEG signals is feasible and shows promising results \cite{du2023decoding} \cite{song2023decoding}. 
This advancement creates new prospects for a wide range of practical applications. 
Specifically, accurately decoding perceptual stimuli from EEG not only deepens our understanding of how the brain processes visual information \cite{li2024visual} but also enables the direct conversion of visual perception or imagined semantics into control commands. 
This will empower emerging BCI technologies, assisting individuals with disabilities in achieving efficient interaction with external devices or facilitating neurorehabilitation training \cite{liu2025recent}. 
Furthermore, this technology holds the potential to expand into cutting-edge fields such as augmented reality \cite{prapas2024connecting} and immersive human-computer interaction \cite{blu2025towards}.

Traditionally, the decoding process of EEG involves learning a mapping function \cite{jin2024mocnn}, \cite{du2022product}, \cite{song2022eeg}, \cite{10197212}, \cite{horikawa2017generic} to associate extracted neural activity features with predefined class labels to achieve the recognition of specific patterns.
However, in visual stimulus-evoked EEG decoding tasks, the scarcity of neural data and the lack of effective training guidance \cite{jiao2019decoding} make it difficult to learn generalizable semantic representations from complex neural signals.
This fundamental limitation leaves traditional approaches severely challenged in accurately decoding visual neural signals, alleviating overfitting, and generalizing toward unseen categories \cite{ahmed2021object}.
In recent research on decoding visual neural representations, researchers have begun incorporating multimodal data \cite{palazzo2020decoding}, \cite{jeong2022real}, \cite{de2024cross}, such as images or text \cite{du2023decoding}, \cite{song2023decoding}.
This aims to mitigate the scarcity of brain activity data and to align visual neural signals with other modalities for decoding or reconstruction \cite{li2024visual}, \cite{bai2023dreamdiffusion}. 
Pioneeringly, Du \textit{et al.} \cite{du2023decoding} proposed BraVL, a trimodal Brain-Visual-Linguistic representation learning framework that aligns neural, visual, and linguistic signals within a shared latent space via deep generative modeling and intra-/inter-modality mutual information regularization \cite{sutter2021generalized}.
This work provides early evidence for the practical feasibility of decoding novel visual categories from human brain activity through multimodal latent alignment.
Nevertheless, under limited neural data, the decoding gains of such generative alignment remain modest for EEG signals.

\begin{figure*}[ht]
\centering
\includegraphics[width=1.0\linewidth]{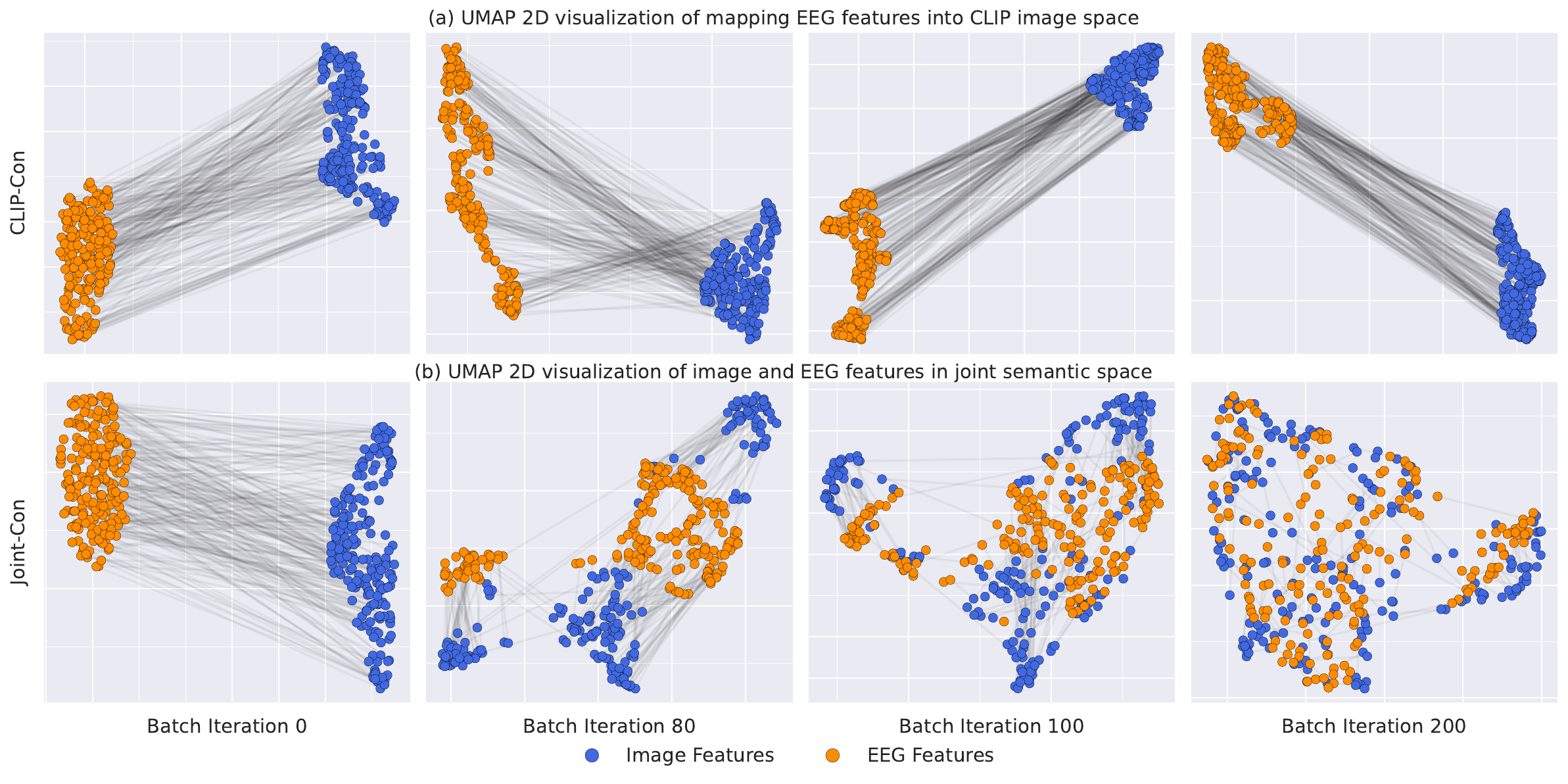}
\caption{2D UMAP \cite{mcinnes2018umap} visualization of 200 test Image-EEG pairs for Subject 8 across different training iterations. (a) CLIP-Con: Directly mapping EEG features into the CLIP image space; the modality gap persists as training progresses. (b) Joint-Con: Within our constructed Joint Semantic Space, the modality gap among test samples gradually vanishes throughout the training process.}
\label{fig:ebshow}
\end{figure*}

Recently, discriminative multimodal alignment \cite{radford2021learning}, \cite{khosla2020supervised} methods have demonstrated promising performance in EEG-based visual decoding. 
These methods typically map extracted EEG features into a pre-trained multimodal embedding space to align with semantic priors and subsequently perform decoding via metric learning.
NICE \cite{song2023decoding} employs a self-supervised framework that uses multimodal contrastive learning (MCL) to align EEG features with CLIP \cite{radford2021learning} image embeddings.
This end-to-end training approach achieves outstanding decoding performance.
Li \textit{et al.} \cite{li2024visual} proposed an effective EEG encoder called ATM, which can extract spatio-temporal representations of EEG signals, further matching them with image embeddings. 
However, as illustrated in Fig. \ref{fig:ebshow}(a), these methods demonstrate that after applying MCL, the embeddings of EEG samples and those of the CLIP image are located in completely separate regions on a hypersphere.
This phenomenon is known as a \textit{modality gap} \cite{liang2022mind}, which creates challenges in accurately cross-modal matching, leading to a decline in model performance on classification or retrieval tasks.
Additionally, although the visual or textual features used for alignment are typically extracted from pre-trained models, not all feature dimensions correspond to the high-level semantic information involved in object recognition as reflected in EEG signals. 
Under fixed prior representations, directly enforcing cross-modal similarity on holistic features may lead the model to align task-irrelevant components, thereby weakening the semantic consistency between modality representations.
As a result, the model may rely on poorly generalizable representations or overfit to noise, particularly for low signal-to-noise ratio signals such as EEG.

To address the aforementioned challenges, we construct a Visual–EEG Joint Semantic Space to bridge the gap between visual and neural representations.
By projecting features from both modalities into this shared latent space, the distributional discrepancies inherent in the prior feature spaces are effectively mitigated.
As illustrated in Fig. \ref{fig:ebshow}(b), the proposed joint semantic space progressively eliminates the modality gap as training process, thereby facilitating more precise cross-modal alignment.
Building on this, we propose two novel approaches to further enhance the semantic consistency of cross-modal representations.
First, we propose the Visual-EEG Semantic Decoupling Network (VE-SDN) to explicitly disentangle semantic-related components within modality representations and dedicate the alignment process solely to these semantic features. 
Meanwhile, a cyclic consistency reconstruction mechanism strictly enforces their semantic consistency.
Quantitative analysis indicates that the proposed VE-SDN facilitates better cross-modal representation alignment, resulting in consistent gains in EEG decoding accuracy.
In addition, inspired by the invariance of core object recognition in the brain \cite{dicarlo2012does}, where visually diverse instances from the same category can still evoke stable high-level semantic representations, we propose a Neural-Guided Intra-Class Consistency (NGIC) objective. 
This approach ensures that distinct visual samples of the same class maintain a consistent alignment distance with their corresponding neural anchors during MCL, which in turn further enhances the robustness of visual representations and improves the final decoding performance.

In summary, the contributions of this work are as follows:
\begin{enumerate}
\item We construct a Visual-EEG Joint Semantic Space to minimize the distributional discrepancies stemming from direct mapping into a prior space, effectively alleviating the modality gap between representations.
\item We propose a Visual-EEG Semantic Decoupling Network to further strengthen the semantic consistency between modality representations. Quantitative analyses of alignment metrics and mutual information confirm that VE-SDN improves cross-modal alignment, proving highly beneficial for the final EEG decoding performance.
\item Drawing on insights from the mechanisms of visual object understanding, we propose a Neural-Guided Intra-Class Consistency approach, further improving visual representation robustness and EEG decoding accuracy.
\item Extensive experiments on large-scale visual-EEG data validate the effectiveness of our proposed method. Compared to strong baselines, our approach improves EEG decoding accuracy by a large margin in both intra- and inter-subject settings, providing feasibility for potential practical applications.
\end{enumerate}

The remainder of this article is organized as follows. 
Section \ref{sec:related_work} reviews related work on EEG-based visual decoding and discriminative multimodal contrastive learning. 
Section \ref{sec:proposed_method} formulates the zero-shot EEG visual decoding task and details the proposed method, including its training and inference pipelines. 
Section \ref{sec:experiments} outlines the experimental setup, data preprocessing, and implementation details. 
Section \ref{sec:experimental_results} presents comprehensive experimental results, followed by an in-depth analysis of the proposed framework in Section \ref{sec:analysis}. 
Finally, Section \ref{sec:dis_lim_con} discusses limitations, suggests future research directions, and concludes this article.
 
\begin{figure*}[ht]
    \centering
    \includegraphics[scale=1.05]{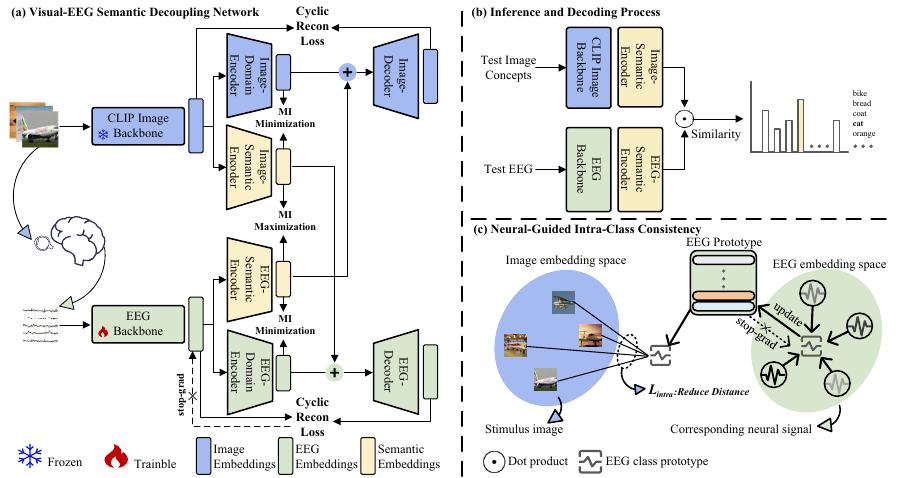}
    \caption{The overview of the proposed Visual-EEG Semantic Decoupling Network.
    (a) The main pipeline for semantic decoupling of stimulus images and EEG signals. 
    (b) The inference and decoding process of our proposed VE-SDN.
    (c) Neural-Guided Intra-Class Consistency constraint in the process of the alignment of semantic-related features. }
    \label{fig:ve_sid}
\end{figure*}

\section{Related Work}
\label{sec:related_work}
\subsection{EEG-Based Visual Neural Decoding}
Owing to the high temporal resolution, cost-effectiveness, and portability of EEG, the decoding of visual-evoked EEG signals is increasingly emphasized in research \cite{zhang2024convolutional}, \cite{ko2024eeg}, \cite{song2023decoding}, \cite{wang2023eeg}, \cite{wang2023ssgcnet}.
Early decoding of semantic categories from EEG signals evoked by natural images involved solely a limited number of classes \cite{kaneshiro2015representational}. 
Spampinato \textit{et al.} \cite{spampinato2017deep} record EEG signals for 40 classes of visual stimuli and propose an LSTM-based method that achieves an accuracy of 82.9\% in classification tasks.
Then, Palazzo \textit{et al.} \cite{palazzo2020decoding} further improve this classification result and introduce a multimodal learning approach to perform classification and saliency detection on visual images and EEG signals.
However, the above block-based experimental paradigm design leads to flaws in the data itself \cite{li2020perils}, resulting in the classification of temporal artifacts rather than stimulus-related activity.
Recent research collected a large-scale visual-EEG dataset \cite{gifford2022large}, \cite{hebart2019things} with 16,740 image stimuli of 1,854 concepts using the rapid serial visual presentation (RSVP) paradigm \cite{won2022eeg}, \cite{lees2018review}. 
This work \cite{gifford2022large} adopts a rapid-event design to avoid the effects of EEG temporal autocorrelation and demonstrates the distinguishability between EEG visual categories.
Building on this work, Du \textit{et al.} \cite{du2023decoding} propose a Brain-Visual-Linguistic trimodal representation learning framework, demonstrating that decoding novel visual categories from human brain activities is practically feasible. 
Song \textit{et al.} \cite{song2023decoding} propose a multimodal contrastive learning approach that improves the accuracy of zero-shot neural decoding for EEG.
Due to the weaknesses of the EEG encoder, Li \textit{et al.} \cite{li2024visual} propose a powerful EEG encoder named ATM, which effectively aligns EEG with image embeddings and further improves the accuracy of EEG visual decoding.
In this work, we explore the semantic consistency between visual information and neural signals and effectively align them in a joint semantic space through the explicit extraction of semantic-related parts to achieve better decoding performance.

\subsection{Multimodal Contrastive Learning}
Multimodal Contrastive Learning aims to maximize the similarity between matched multimodal pairs and minimize the similarity between all the mismatched pairs. 
This alignment method enables the learned representations to excel in multiple downstream tasks and demonstrates strong zero-shot classification capabilities \cite{radford2021learning} \cite{jia2021scaling}.
Recent research indicates a phenomenon known as the \textit{modality gap} \cite{liang2022mind} during multimodal contrastive learning.
This modality gap refers to the representations from different modalities tend to reside in distinct regions of a joint representation space \cite{xu2021videoclip}, \cite{zhang2022contrastive}. 
Previous studies \cite{liang2022mind} demonstrate that reducing this gap may improve model performance in zero-shot classification tasks.
However, recent analyses suggest that gap reduction alone may be insufficient, as \cite{jiang2023understanding} shows that reducing the modality gap generally does not yield optimal results for downstream predictive tasks and instead advocates constructing a meaningful latent modality structure.
For instance, CyCLIP \cite{goel2022cyclip} constructs a geometric constraint in multimodal representation spaces with cyclic consistency to improve performance on downstream predictive tasks. 
In this work, alongside MCL, we construct stable class-specific neural semantic anchors to progressively minimize the distance between visual samples and their corresponding EEG anchors.
Inspired by the invariance property of core object recognition in the human brain, we impose a direct distance constraint that reduces intra-class variability and further improves EEG decoding performance.

\section{Proposed Method}
\label{sec:proposed_method}
In this section, we present a comprehensive description of the proposed method.
We formulate the zero-shot neural decoding task for visual object recognition and introduce the overall architecture of the proposed VE-SDN and its key components, as illustrated in Fig. \ref{fig:ve_sid}.
We then elaborate on the Neural-Guided Intra-Class Consistency approach.
Finally, we describe the training strategy and inference procedure.

\subsection{Problem Definition: Zero-Shot Neural Decoding}
The task setup follows \cite{song2023decoding}, employing only visual~stimuli images and EEG pairs without introducing additional data. 
Specifically, given a seen class dataset $ D^{seen} =  \left \{ \left ( x_{b}, x_{v}, y | x_{b}\in X_{b}, x_{v}\in X_{v}, y\in Y  \right )  \right \} $, where $ X_{b} $ and $ X_{v} $ denote the EEG signals and stimulus images, and $ Y $ denotes the corresponding class labels. 
The seen class labels span from $ 1 $ to $ S $, $ y \in L = \left \{1,...,S \right \} $. 
Given an unseen class dataset $ D^{unseen} = \left \{ \left ( x_{b}^{u},x_{v}^{u}, y^{u} | x_{b}^{u}\in X_{b}^{u}, x_{v}^{u}\in X_{v}^{u}, y^{u}\in Y^{u}  \right )  \right \}  $, where  $ X_{b}^{u}, X_{v}^{u}, Y^{u} $ represents the set of unseen class EEG, images, and class labels, respectively. 
The unseen class labels span from $ S+1 $ to $ S+U $, $ y^{u} \in L^{u} = \left \{S+1,...,S+U \right \} $. 
The seen and unseen classes are mutually exclusive, i.e., $ L\cap L^{u} = \emptyset $. 
The unseen class features $ X_{b}^{u}, X_{v}^{u} $ are not available during training. 
At the testing phase, $ X_{v}^{u} $ serves as novel concepts for new visual categories, with the decoding of EEG features $ X_{b}^{u} $ achieved by calculating their similarity to these visual concepts.

\subsection{Cross-Modal Semantic Information Decoupling}
\textit{1) Semantic and Domain Feature Extraction:}
The total pipeline of our model VE-SDN is illustrated in Fig. \ref{fig:ve_sid} (a). 
We employ a fixed pre-trained CLIP Vision Transformer (ViT) as the visual backbone and a trainable ATM-S \cite{li2024visual} EEG encoder as the neural backbone to extract feature representations from raw stimulus images and preprocessed EEG signals.
Specifically, given $ N $ paired image-EEG samples, the visual backbone produces image embeddings $h_v \in \mathbb{R}^{N \times 1024}$, while the EEG backbone extracts EEG embeddings $ h_{b} = F_{b}(x_{b}) \in \mathbb{R}^{N \times 1024} $ from the neural signals $ x_{b} \in \mathbb{R}^{N \times C \times T} $, where $ F_{b}, C, T $ represents the EEG backbone, the channel and the time point of EEG signals, respectively. 
The extracted image and EEG embeddings are then fed into their respective modality-specific semantic and domain encoders, which are denoted as:
\begin{equation}
\label{eq:extract_feas}
\begin{aligned}
    z_v^{s} &= \Phi_v(h_v) \qquad
    z_v^{d} = \Psi_v(h_v)  \\
    z_b^{s} &= \Phi_b(h_b) \qquad
    z_b^{d} = \Psi_b(h_b)
\end{aligned}
\end{equation}
where $z_{v}^{s}, z_{v}^{d}, z_{b}^{s}, z_{b}^{d}$ represent the semantic and domain features of images and EEG embeddings, respectively. $ \Phi, \Psi $ are the semantic and domain encoder networks for each modality.

We aim to explicitly extract semantic-related information from image and EEG embeddings, enabling the extracted features to capture modality-invariant and generalizable components.
In contrast, domain features characterize modality-specific factors and are disentangled from the semantic representations.
Inspired by prior studies on domain adaptation \cite{xia2023achieving}, \cite{cheng2020club} and disentangled representation learning \cite{wang2024disentangled}, we adopt a mutual information maximization-minimization adversarial strategy to progressively decouple the semantic and domain components of each modality during training, as follows.

\textit{2) Mutual Information Minimization:}
Assuming that the semantic and domain parts of each modality are uncorrelated, information decoupling can be achieved by minimizing an upper bound on the mutual information between the corresponding semantic and domain features. 
To this end, we employ CLUB \cite{cheng2020club} to estimate the upper bound of mutual information between features, defined as:
\begin{equation}
\begin{split}
I_{CLUB}(z_{m}^{d};z_{m}^{s}) & \mathrel{:=} \mathbb{E}_{p(z_{m}^{d},z_{m}^{s})}\big[\log p(z_{m}^{s}|z_{m}^{d})\big] \\
& - \mathbb{E}_{p(z_{m}^{d})p(z_{m}^{s})}\big[\log p(z_{m}^{s}|z_{m}^{d})\big]
\end{split}
\end{equation}
where $ I(z_{m}^{d};z_{m}^{s}), m \in \{ v,b \} $ are the mutual information between the two modalities semantic and domain features. 
Since the conditional distribution $ p(z^{s} | z^{d}) $ is intractable in practice, we employ a variational inference approach, parameterizing $ p(z^{s} | z^{d}) $ with a neural network $ \theta $, and to approximate the upper bound of mutual information with samples $ \{ (z_{m,i }^{d}, z_{m,i}^{s}) \}_{i=1}^{N} $, formulated as:
\begin{equation}
\begin{split}
\hat{I}_{vCLUB}(z_{m}^{d};z_{m}^{s})  & = \frac{1}{N}\sum\nolimits_{i=1}^{N} \big[\log q_{\theta _{m}}(z_{m,i}^{s}|z_{m,i}^{d}) \\
& - \frac{1}{N}\sum\nolimits_{j=1}^{N} \log q_{\theta _{m}}(z_{m,j}^{s}|z_{m,i}^{d})\big]
\end{split}
\end{equation}
According to \cite{cheng2020club} Theorem 3.2, we need to minimize the discrepancy between the joint distribution $ p(z^{d}, z^{s}) $ and the variational joint distribution $ q_{\theta}(z^{d}, z^{s}) $, to ensure that $ \hat{I}_{vCLUB}(z^{d};z^{s}) $ still represents a reliable upper bound of mutual information. 
Therefore, we adopt the KL divergence to measure the difference between two distributions, formulated as:
\begin{equation}
\label{eq:minkl}
\begin{split}
& \min_{\theta}KL(p(z^{d},z^{s})||q_{\theta}(z^{d}, z^{s})) = \\
& \min_{\theta}\mathbb{E}_{p(z^{d},z^{s})}[\log p(z^{s}|z^{d})] - \mathbb{E}_{p(z^{d},z^{s})}[\log q_{\theta}(z^{s}|z^{d})] 
 \end{split}
\end{equation}
In Eq. (\ref{eq:minkl}), the first term has no relation with parameter $\theta $, thus to minimize the $ KL(p(z^{d}, z^{s})||q_{\theta}(z^{d}, z^{s})) $ can be achieved by minimizing the negative log-likelihood with samples $ \{ (z_{m,i}^{d}, z_{m,i}^{s}) \}_{i=1}^{N} $, as follows:
\begin{equation}
\label{eq:loglikeli}
L(\theta) = -\frac{1}{N} { \sum\nolimits_{i=1}^{N}} \log q_{\theta}(z^{s}|z^{d})  
\end{equation}
which is the unbiased estimation of $ \mathbb{E}_{p(z^{d},z^{s})}[\log q_{\theta}(z^{s}|z^{d})] $.
During training, we decouple the semantic and domain representations by minimizing the estimated mutual information between them for both image and EEG embeddings, as follows:
\begin{equation}
\label{eq:mi_min}
    L_{MI} = \hat{I}(z_{v}^{d};z_{v}^{s})  + \hat{I}(z_{b}^{d};z_{b}^{s}) 
\end{equation}
The approximate neural networks $ \theta $ for the two modalities and the main network are trained alternately during the optimization process.

\textit{4) Mutual Information Maximization:}
To determine the cross-modal semantic consistency features between the extracted image and EEG embeddings.
We aim to maximize the mutual information between semantic-related parts $ z_{v}^{s} $ and $ z_{b}^{s} $. 
Following the approach proposed in \cite{oord2018representation}, our goal is to maximize the lower bound on mutual information by minimizing the InfoNCE loss, formulated as: 
\begin{equation}
\begin{split}
\label{eq:con}
L_{con} = & -\frac{1}{2N} \sum_{i=1}^{N} \biggl[ \log\frac{exp(f(z_{v,i}^{s},z_{b,i}^{s})/\tau )}{\sum_{j=1}^{N}exp(f(z_{v,i}^{s},z_{b,j}^{s})/\tau)} \\
& + \log\frac{exp(f(z_{b,i}^{s},z_{v,i}^{s})/\tau )}{\sum_{j=1}^{N}exp(f(z_{b,i}^{s},z_{v,j}^{s})/\tau)} \biggr] 
\end{split}
\end{equation}
where $ f( \cdot , \cdot) $ represents the cosine similarity, and $ \tau $ represents the learnable temperature parameter.
We explicitly decouple semantic representations of image and EEG modalities via an adversarial mutual information maximization–minimization strategy, while aligning cross-modal semantics through MCL.

\textit{4) Inter-modality Semantic Consistency:} 
To avoid the degenerate case of modality domain features, and to further enhance the semantic consistency of visual and EEG semantic features. 
We employ cyclic consistency signal reconstruction by reconstructing the original visual image embeddings using visual domain features combined with the EEG semantic features and vice versa, formulated as:
\begin{equation}
\begin{split}
\label{eq:cyclic_recon}
L_{recon}& =\frac{1}{2}(MSE(h_{v},D_{v}(z_{v}^{d},z_{b}^{s})) \\
& + MSE(h_{b},D_{b}(z_{b}^{d},z_{v}^{s})))  
\end{split}
\end{equation}
where $MSE(\cdot, \cdot)$ denotes the mean squared loss, and $D_{v}(\cdot , \cdot)$, $D_{b}(\cdot, \cdot) $ denote the decoder networks for the visual and EEG modalities, respectively.
In this work, the decoder takes as input the concatenation of features from the two modalities.
By enforcing semantic alignment via cyclic consistency reconstruction, the representations of the two modalities are forced to contain consistent information.

\begin{algorithm}[ht]
\caption{Training Algorithm of the VE-SDN}
\label{alg:algorithm1}
\KwIn{Training Image-EEG pairs $D^{seen}$; Image features $h_{v}$; Temperature $\tau$; Hyper-parameters $\lambda_{1}, \lambda_{2}, \lambda_{3}, \alpha$.}
\KwOut{EEG backbone $F_{b}$; EEG-semantic encoder $\Phi_{b}$; Image-semantic encoder $\Phi_{v}$.}
\BlankLine
\textbf{Initialization:} Initialize parameters $\{ F_{b}, \Phi_{b}, \Psi_{b}, \theta_{b}, D_{b}, \Phi_{v}, \Psi_{v}, \theta_{v}, D_{v} \}$ and prototype bank $C_{b}$; \\
\For{each training iteration}{
    \For{each sampled mini-batch $\{ h_{v,i}, x_{b,i}, y_{i} \}_{i = 1}^{N}$}{
        Extract EEG features $h_{b} \gets F_{b}(x_{b})$; \\
        Extract semantic $z^{s}$ and domain $z^{d}$ features for both modalities via Eq.~(\ref{eq:extract_feas}); \\
        Apply $\ell_2$-normalization: $z \gets z / \|z\|_2, \ \forall z \in \{ z_{v}^{s}, z_{v}^{d}, z_{b}^{s}, z_{b}^{d} \}$; \\
        Compute class-wise mean $\bar{z}_{b}^{s}$ via Eq.~(\ref{eq:cal_average_zbs}); \\
        Update EEG prototypes $C_{b} \gets \alpha C_{b} + (1-\alpha)\bar{z}_{b}^{s}$ via Eq.~(\ref{eq:momentum_update}); \\
        
        \For{$k \gets 1$ \KwTo $n_{logli}$}{
            Evaluate Log-likelihood $L(\theta_{v}), L(\theta_{b})$ given $(z^{d}, z^{s})$ via Eq.~(\ref{eq:loglikeli}); \\
            Update variational networks $\theta_{v}, \theta_{b}$ by maximizing $L(\theta_{v}), L(\theta_{b})$; \\
        }
        
        Compute MI minimization loss $L_{MI} \gets \hat{I}(z_{v}^{d}; z_{v}^{s}) + \hat{I}(z_{b}^{d}; z_{b}^{s})$ via Eq.~(\ref{eq:mi_min}); \\
        Compute InfoNCE loss $L_{con} \gets \mathcal{L}_{NCE}(z_{v}^{s}, z_{b}^{s})$ via Eq.~(\ref{eq:con}); \\
        Compute cyclic recon loss $L_{recon} \gets \mathcal{L}_{MSE}^{v} + \mathcal{L}_{MSE}^{b}$ via Eq.~(\ref{eq:cyclic_recon}); \\
        Compute intra-class loss $L_{intra} \gets \frac{1}{N} \sum \bar{d}_{k}$ via Eqs.~(\ref{eq:average_distance})-(\ref{eq:loss_intra}); \\
        Compute total objective $L \gets L_{con} + \lambda_{1}L_{MI} + \lambda_{2}L_{recon} + \lambda_{3}L_{intra}$; \\
        Update main network parameters by minimizing $L$; \\
    }
}
\end{algorithm}

\subsection{Neural-Guided Intra-Class Consistency}
We aim to leverage label information to improve model performance. 
However, directly pulling cross-modal samples of the same category closer using label supervision (e.g., Supervised Contrastive Learning \cite{khosla2020supervised}) does not lead to performance gains, as shown in the experimental results in Section \ref{sec:exp_on_supcon}.
Hence, we design a brain-aligned Neural-Guided Intra-Class Consistency objective.
As \cite{dicarlo2012does} proposes, although the same object may elicit diverse retinal response patterns, the visual system can establish an invariant, concept-level representation.
Accordingly, in the joint semantic space, we aim to establish this consistency by aligning visual representations of the same category to their corresponding neural semantic anchors, thereby minimizing the intra-class variability of visual representations.
Specifically, we adopt an asymmetric cross-modal anchoring strategy, in which an EEG prototype is maintained during training as a stable class-level neural semantic anchor, and image embeddings are aligned to these neural anchors.
We randomly initialize the EEG prototypes for all $ K $ categories as $ C_{b} = \{ c_{1}, c_{2},..., c_{k}\} $.
During each training iteration, we compute the class-wise mean $ \bar{z}_{b,k}^{s} $ of EEG semantic features within the mini-batch, as follows:
\begin{equation}
\label{eq:cal_average_zbs}
\bar{z}_{b,k}^{s} = \frac{1}{ | Y_{b,k}| } \sum_{ z_{b}^{s} \in Y_{b,k}}z_{b}^{s}
\end{equation}
where $ Y_{b,k} $ denotes EEG semantic features belonging to class $ k $ in the current mini-batch, and $ |Y_{b,k}| $ indicates its cardinality.

\textit{1) Prototype Momentum Update:}
We maintain a memory bank to store EEG semantic prototypes and update them dynamically using an exponential moving average (EMA) strategy.
After computing the class-wise centroid of EEG semantic features within each mini-batch, the prototype set \( C_b \) is updated as
\begin{equation}
\label{eq:momentum_update}
c_k \leftarrow \alpha c_k + (1 - \alpha)\,\bar{z}_{b,k}^{s}
\end{equation}
where \( \alpha \) denotes the momentum coefficient.

\textit{2) Intra-class Consistency Loss:}
The visual system can unify different response patterns to the same object, thereby enabling the brain to identify and retrieve these visual samples precisely. 
Similarly, we enforce this consistency in visual representations by explicitly minimizing the distance between diverse visual samples of the same category and their corresponding EEG prototypes.
Concretely, for a mini-batch visual semantic samples $ z_{v}^{s} $, we calculate the distance between each sample and its corresponding EEG semantic prototype $ c_{k} $.
Then, we compute the average intra-class distance within each category $ k $:
\begin{equation}
\label{eq:average_distance}
\bar{d}_{k} = \frac{1}{ | Y_{v,k} | } \sum_{z_{v}^{s} \in Y_{v,k}}d(z_{v}^{s}, c_{k} )
\end{equation}
where $ d(\cdot , \cdot) $ denotes the cosine distance, consistent with the metric used in the main MCL objective.
$ \bar{d}_{k} $ represents the average distance from intra-class visual samples to the neural prototype, and $ Y_{v,k} $ is the set of visual semantic features with label $ k $.
The intra-class consistency loss is as follows:
\begin{equation}
\label{eq:loss_intra}
L_{intra} =  \frac{1}{ N } \sum_{k = 1}^{N} \bar{d}_{k}
\end{equation}
where $ N $ denotes the number of categories present in a mini-batch.
During training, the intra-class consistency loss progressively aligns visual semantic features towards the EEG neural semantic anchors. 
Although the inherent modality gap precludes a perfect cross-modal alignment (a vanishing cosine distance), minimizing these distance discrepancies concurrently suppresses intra-class variability. 
This process effectively shapes a robust latent structure in the embedding space that is conducive to EEG signal decoding.

It is worth noting that the implementation of our Neural-Guided Intra-Class Consistency is closely related to Center Loss \cite{wen2016discriminative} or Cross-Modal Center Loss \cite{jing2021cross}.
In Section. \ref{sec:intra_class}, we provide a detailed analysis to empirically demonstrate that our proposed asymmetric cross-modal anchoring strategy is more effective for EEG decoding tasks.

\subsection{Training and Inference}
Algorithm \ref{alg:algorithm1} shows the pseudocode of model training. In the training process, there are two approximate networks $ \theta_{b}, \theta_{v} $ alternating optimization with the main network, and the loss denoted as:
\begin{equation}
L_{loglikei} = L(\theta_{b}) + L(\theta_{v})
\end{equation}
The overall loss for our main network is defined as :
\begin{equation}
L = L_{con} + \lambda_{1}L_{MI} + \lambda_{2}L_{recon} + \lambda_{3}L_{intra}
\end{equation}
where $ \lambda_{1}, \lambda_{2} $ and $ \lambda_{3} $ are hyperparamers.

In the decoding process, we use the visual image as concept templates. Then, input the test EEG signals and visual templates into the model to get the semantic-related features. We get the results by calculating the cosine similarity between test EEG features and visual templates, which is denoted as:
\begin{equation}
P_{zsl}(x_{b}^{u}) = \mathop{max}\limits_{j} \langle z_{b}^{s}, z_{v_j}^{s} \rangle_{j=S+1}^{S+U}
\end{equation}

\begin{table*}[!ht]\Huge
\renewcommand{\arraystretch}{1.2} 
\centering
\caption{Ablation study on the 200-way zero-shot classification task, reporting top-1 and top-5 accuracies (\%) for each approaches. The best results are highlighted in \textbf{bold}.}
\label{tab:abla_table}

\resizebox{\linewidth}{!}{
\begin{tabular}{@{} l *{22}{c} @{}} 
\toprule
& \multicolumn{2}{c}{Subject 1} & \multicolumn{2}{c}{Subject 2} & \multicolumn{2}{c}{Subject 3} & \multicolumn{2}{c}{Subject 4} & \multicolumn{2}{c}{Subject 5} & \multicolumn{2}{c}{Subject 6} & \multicolumn{2}{c}{Subject 7} & \multicolumn{2}{c}{Subject 8}  & \multicolumn{2}{c}{Subject 9}  & \multicolumn{2}{c}{Subject 10} & \multicolumn{2}{c}{Avg} \\
\cmidrule(lr){2-3} \cmidrule(lr){4-5} \cmidrule(lr){6-7} \cmidrule(lr){8-9} \cmidrule(lr){10-11} \cmidrule(lr){12-13} \cmidrule(lr){14-15} \cmidrule(lr){16-17} \cmidrule(lr){18-19} \cmidrule(lr){20-21} \cmidrule(lr){22-23}     
Method  & top-1 & top-5 & top-1 & top-5 & top-1 & top-5 & top-1 & top-5 & top-1 & top-5  & top-1 & top-5  & top-1 & top-5  & top-1 & top-5 & top-1 & top-5  & top-1 & top-5  & top-1 & top-5  \\

\midrule 
\multicolumn{23}{c}{\textbf{Intra-subject:} train and test on one subject} \\
\midrule 

CLIP-Con    & 23.7 & 53.0 & 24.2 & 54.6 & 29.1 & 62.0 & 27.4 & 63.7 & 21.6 & 49.0 & 25.8 & 63.1 & 25.0 & 54.9 & 37.2 & 68.0 & 26.1 & 58.3 & 28.0 & 62.3 & 26.81 & 58.89 \\
Joint-Con  &  33.8 & 65.5 & 32.0 & 63.6 & 41.8 & 73.8 & 39.3 & 73.3 & 28.5 & 60.9 & 37.5 & 72.5 & 34.7 & 66.1 & 50.7 & 82.6 & 35.9 & 67.2 & 46.0 & 79.2 & 38.02 & 70.47 \\
Joint-Con+NGIC & \textbf{40.2} & {70.9} & 34.6 & \textbf{68.7} & {42.8} & 74.2 & 41.0 & 73.6 & \textbf{30.9} & \textbf{63.1} & 40.3 & {75.0} & {36.8} & {69.5} & {53.5} & \textbf{84.0} & {38.7} & {71.4} & \textbf{50.2} & \textbf{81.1} & {40.90} & {73.15} \\
VE-SDN    & 34.0 & 66.7 & {34.9} & 64.6 & 41.7 & {74.3} & {43.3} & {74.8} & 29.4 & 59.5 & \textbf{42.6} & 74.1 & 34.9 & 67.7 & 52.2 & 81.5 & 38.1 & 70.7 & 45.3 & 78.4 & 39.64 & 71.23 \\
VE-SDN+NGIC  & {39.4} & \textbf{71.8} & \textbf{35.6} & \textbf{68.7} & \textbf{45.0} & \textbf{74.4} & \textbf{44.9} & \textbf{75.5} & {30.6} & {62.5} & {42.2} & \textbf{76.7} & \textbf{36.9} & \textbf{69.8} & \textbf{54.3} & {83.4} & \textbf{40.2} & \textbf{73.6} & {48.4} & {80.7} & \textbf{41.75} & \textbf{73.71} \\

\midrule 
\multicolumn{23}{c}{\textbf{Inter-subject:} leave one subject out for test} \\
\midrule 

CLIP-Con  & 10.4 & 30.3 & 12.5 & 33.9 & \textbf{9.8} & \textbf{33.3} & 11.3 & 35.6 & 9.0 & 28.8 & 8.8 & 30.8 & 5.8 & 24.4 & 11.8 & 31.8 & 12.9 & 38.8 & 15.8 & 40.9 & 10.81 & 32.86  \\
Joint-Con  &  {12.0} & 31.0 & {16.8} & 39.8 & {9.1} & {30.3} & \textbf{15.2} & {38.9} & 13.4 & {34.7} & {12.7} & \textbf{35.6} & 8.6 & 27.0 & 12.6 & 35.9 & 14.6 & 40.8 & 17.1 & 44.1 & 13.21 & 35.81 \\
Joint-Con+NGIC  & {12.0} & 30.5 & 16.2 & 38.7 & {9.1} & 29.3 & {14.3} & 37.1 & \textbf{13.8} & {35.2} & {11.8} & 35.0 & {9.9} & 26.4 & \textbf{13.1} & \textbf{38.6} & 13.9 & 41.9 & 18.9 & 45.7 & {13.30} & 35.84 \\
VE-SDN  & \textbf{12.4} & 33.3 & \textbf{17.9} & \textbf{42.3} & 8.0 & 31.6 & 14.6 & \textbf{41.9} & 12.3 & 36.1 & \textbf{13.4} & 33.5 & 9.8 & \textbf{29.9} & \textbf{13.1} & 38.5 & 15.4 & \textbf{42.4} & 20.0 & 45.9 & \textbf{13.69} & \textbf{37.54} \\
VE-SDN+NGIC & 11.2 & \textbf{34.2} & 15.0 & 40.2 & 9.6 & 29.7 & 14.3 & 40.5 & 12.9 & \textbf{36.6} & 11.6 & 33.2 & \textbf{10.3} & 28.6 & 11.9 & 38.1 & \textbf{16.5} & 42.3 & \textbf{22.2} & \textbf{46.7} & 13.55 & 37.01 \\
\bottomrule

\end{tabular}}
\end{table*}

\begin{figure*}[!ht]
    \centering
    \includegraphics[width=1.0\linewidth]{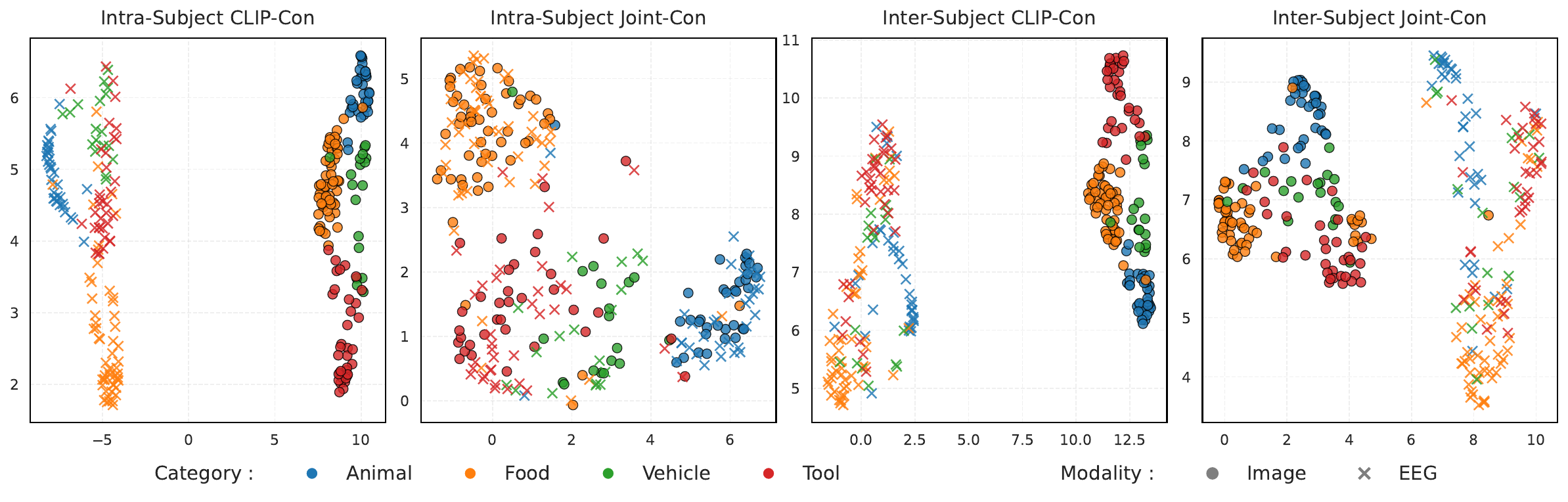}
    \makebox[0.24\linewidth][c]{\hspace{0.0cm}(a)\hspace{-0.0cm}}%
    \makebox[0.24\linewidth][c]{\hspace{0.1cm}(b)\hspace{-0.1cm}}
    \makebox[0.24\linewidth][c]{\hspace{0.2cm}(c)\hspace{-0.2cm}}
    \makebox[0.24\linewidth][c]{\hspace{0.4cm}(d)\hspace{-0.4cm}}%

    \caption{2D UMAP visualization of test image and EEG embeddings under the CLIP-Con and Joint-Con settings for subject 8 in both intra- and inter-subject evaluations. Four representative categories—animals, food, vehicles, and tools—are displayed.}
    \label{fig:con_show}
\end{figure*}

\section{EXPERIMENTS}
\label{sec:experiments}
\subsection{Datasets and Preprocessing}
We evaluate our approach on THINGS-EEG2 \cite{gifford2022large}, a large-scale EEG dataset collected under the RSVP paradigm from 10 human subjects.
The EEG data are collected using 64-channel EASYCAP equipment. 
The training set includes 1654 image classes with each class 10 images, and each image presents 4 times (1654 concepts $\times$ 10 images $\times$ 4 repetitions) per subject. 
The test set includes 200 image classes with each class only 1 image, and each image presents 80 times (200 concepts $\times$ 1 image $\times$ 80 repetitions) per subject. 
In this work, for EEG preprocessing, we adopt the same method as \cite{song2023decoding}. 
The EEG data are segmented into 0-1000 ms trials post-stimulus onset, with baseline correction using the prior 200 ms average. 
The data retain all electrodes, are downsampled to 250 Hz, and undergo multivariate noise normalization using the training data. 
Also, we averaged each EEG repetition to ensure a high signal-to-noise ratio, yielding 16540 training samples and 200 test samples per subject.
For stimulus images, we utilize fixed pre-trained CLIP-ViT-H-14\footnote{\url{https://huggingface.co/laion/CLIP-ViT-H-14-laion2B-s32B-b79K}} as the image backbone to extract the image features.

\subsection{Experiment Settings}
\textit{1) Evaluation strategy:}
We conduct experiments under both intra-subject and inter-subject settings. For inter-subject evaluation, we adopt a leave-one-subject-out (LOSO) protocol. In each LOSO step, test samples from one subject are used as the test set, while training samples from remaining subjects are used for model training. During training, the checkpoint with the lowest contrastive loss is selected for evaluation. Performance is assessed using Top-1 and Top-5 EEG classification accuracy. Reported results are averaged over five independent runs initialized with different random seeds.

\textit{2) Implementation Details:}
Our experiments are conducted on a single A6000 GPU using PyTorch.
The EEG backbone used in this work is ATM-S \cite{li2024visual}.
The semantic encoder, domain encoder, and modality decoder are implemented as two-layer MLP with GELU activation, with the semantic and domain encoders projecting modality features into a shared 512-dimensional embedding space.
Under the intra-subject setting, the model is trained for 50 epochs, whereas it is trained for 10 epochs in the inter-subject setting, with all other training strategies and hyperparameters kept consistent.
We adopt the AdamW optimizer with a learning rate of 1e-4 and a batch size of 1024. 
Furthermore, the initial temperature parameter for contrastive learning is set to 0.07.
The momentum coefficient $ \alpha $ is initialized to 0.5 and progressively increased to 0.99 using a cosine scheduling strategy. 
Specifically, during the first half of training, only the EEG prototypes are updated via EMA, with the momentum coefficient gradually increasing to stabilize the prototype representations, while the intra-class consistency loss remains inactive. In the second half of training, the intra-class consistency loss is activated to further optimize the model parameters.
The loss weight parameters $ \lambda_{1}, \lambda_{2}, \lambda_{3} $ are set as 1.0, 1.0, 5.0, respectively.

\begin{figure*}[!ht]
\centering
\begin{minipage}{1.0\linewidth}
    \centerline{\includegraphics[width=\linewidth, height=5.1cm]{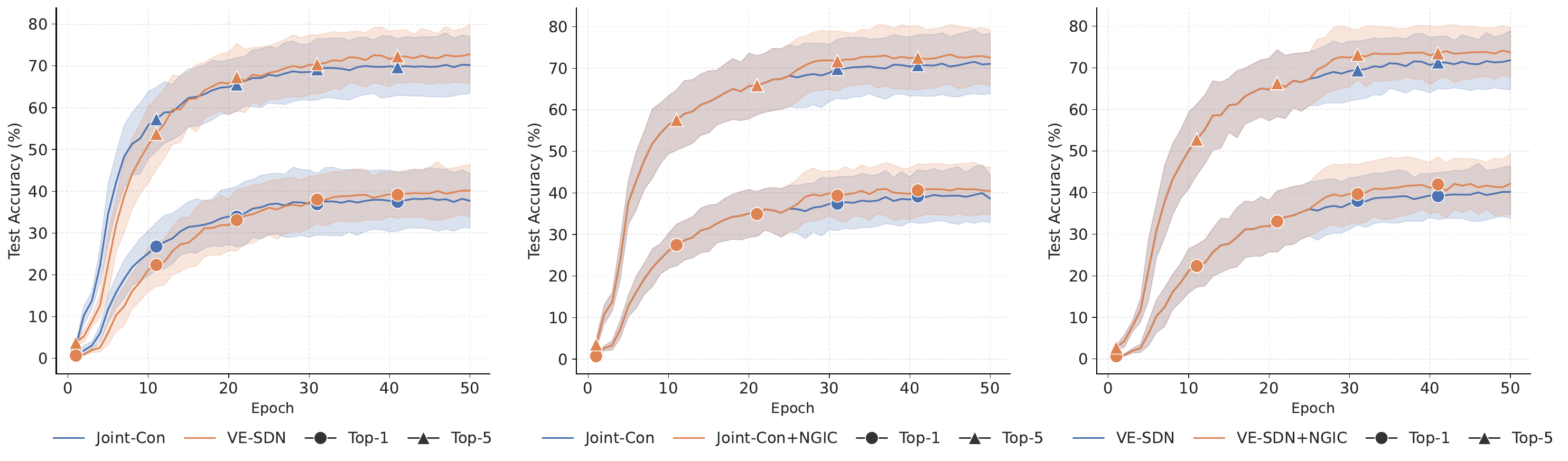}}
    \centerline{(a) Intra-Subject Test Accuracy Curves}
\end{minipage}
\begin{minipage}{1.0\linewidth}
    \centerline{\includegraphics[width=\linewidth, height=5.1cm]{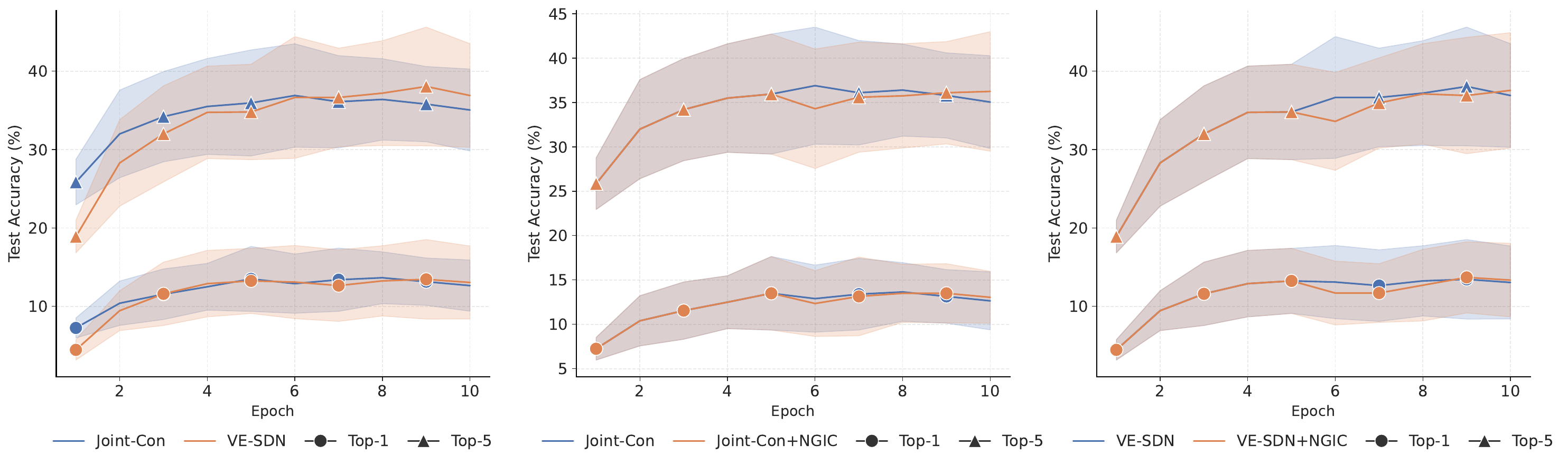}}
    \centerline{(b) Inter-Subject Test Accuracy Curves}
\end{minipage}
\caption{Illustration of the average top-1 and top-5 test accuracy curves across ten subjects under different proposed methods.}
\label{fig:test_acc_curves}
\end{figure*}

\section{Experimental Results}
\label{sec:experimental_results}

\subsection{Ablation Study}
\label{sec:ablation_study}
In this section, we conduct ablation studies on the proposed methods to understand their contributions to overall performance.
We evaluate the proposed methods on the 200-way zero-shot classification task, measuring both Top-1 and Top-5 accuracy.
The results are summarized in Table \ref{tab:abla_table}, and the test accuracy curves of different methods are shown in Fig. \ref{fig:test_acc_curves}.

\textit{1) Ablation on Alignment Space for Contrastive Learning:}
Previous studies directly map EEG embeddings into the CLIP space and performed alignment using contrastive learning. 
However, due to the modality gap, this strategy results in suboptimal alignment between modalities. 
We utilize the same EEG projector as in \cite{li2024visual} (a residual MLP layer) to directly map the extracted EEG embeddings into the CLIP image space while employing a contrastive loss in Eq. (\ref{eq:con}) for alignment.
The average intra- and inter-subject top-1/top-5 accuracy across ten subjects is 26.81/58.89 and 10.81/32.86, respectively, as shown in Table \ref{tab:abla_table} CLIP-Con.
Next, we construct a joint latent space in which both image and EEG embeddings are reprojected into a 512-dimensional joint semantic space via a two-layer MLP with GELU activation, and the same contrastive loss is applied for representation alignment.
The test accuracies are reported in Table \ref{tab:abla_table} Joint-Con. 
The results demonstrate that aligning image and EEG representations within the constructed joint semantic space substantially improves both intra- and inter-subject EEG decoding performance. 
Specifically, the top-1/top-5 accuracies reach 38.02/70.47 for intra-subject evaluation and 13.21/35.81 for inter-subject evaluation, yielding absolute improvements of 11.21/11.58 and 2.40/2.95 over CLIP-Con, corresponding to relative gains of 41.8\%/19.7\% and 22.2\%/9.0\%, respectively.

Figure \ref{fig:con_show} presents the visualization results of test sample embeddings for Subject 8 under both the CLIP-Con and Joint-Con methods. 
Four representative high-level categories were selected from the 200 test classes for illustration. 
It can be observed that, under different experimental settings, intra-semantic class clusters are formed within each modality. 
However, due to the modality gap, as shown in Fig. \ref{fig:con_show} (a) and (c), the CLIP-Con method exhibits distinct distribution regions between modalities in both intra- and inter-subject evaluations, which makes cross-modal sample matching more challenging.

In contrast, Joint-Con reprojects both modalities into a joint semantic space, thereby alleviating the bias introduced by direct mapping. 
Under the intra-subject setting, as shown in Fig. \ref{fig:con_show} (b), this strategy effectively mitigates the modality gap within the visual–EEG joint space.
Consequently, it promotes modality-agnostic intra-semantic class clustering and bridges the distributional gap between visual and EEG samples.
Under the inter-subject Joint-Con setting, as shown in Fig. \ref{fig:con_show} (d), substantial inter-subject variability in EEG signals prevents the constructed joint semantic space, which is learned from non-target subjects, from fully eliminating the distributional gap between the target subject’s EEG and image embeddings. 
Nevertheless, compared with CLIP-Con, the cross-modal distribution discrepancy is partially reduced. 
This observation suggests that the explicitly constructed joint semantic space is more effective in extracting modality-agnostic representations.

\textit{2) Ablation on Explicit Semantic Decoupling:}
To evaluate the effect of explicit semantic decoupling, we incorporate the proposed VE-SDN framework into the joint semantic space. 
This framework employs an MI-based decoupling strategy together with inter-modal semantic consistency to promote the alignment of semantically correlated embeddings across modalities, as detailed in Section \ref{sec:decouple_analysis}. 
As shown in Table \ref{tab:abla_table} and Fig. \ref{fig:test_acc_curves}, VE-SDN consistently outperforms Joint-Con in both intra- and inter-subject settings, yielding top-1/top-5 gains of 1.62/0.76 and 0.48/1.73, respectively. 
These results indicate that explicit semantic decoupling facilitates the learning of more generalizable modality representations, thereby improving zero-shot EEG decoding performance.

\textit{3) Ablation on Intra-class Consistency:} 
We apply Neural-Guided Intra-Class Consistency across different model variants and experimental settings to evaluate its impact on performance.
The results show that the performance gains differ between the intra- and inter-subject settings. Under the intra-subject setting, as illustrated in Fig. \ref{fig:test_acc_curves}(a), activating intra-class consistency regularization leads to clear performance improvements for both Joint-Con and VE-SDN. 
As reported in Table \ref{tab:abla_table}, applying the intra-class loss yields stable performance gains for nearly all subjects. 
Specifically, under the Joint-Con and VE-SDN approaches, the top-1/top-5 accuracies increase by 2.88/2.68 and 2.11/2.48, respectively.

However, under the inter-subject setting, intra-class consistency does not yield stable performance gains and may introduce adverse effects during feature alignment. 
This is because the EEG prototypes are derived from non-target subjects’ embeddings. 
Forcing image embeddings to align with these subject-specific EEG prototypes inevitably introduces a train–test domain shift due to inter-subject variability.
Although we design an explicit semantic decoupling framework, it must be acknowledged that fully disentangling subject-invariant semantic representations from highly subject-specific EEG signals remains a challenging problem. 
Addressing this issue requires additional auxiliary constraints to facilitate deeper decoupling. 
For example, in the proposed VE-SDN framework, additional constraints are imposed to ensure that domain-related features can reliably predict subject identity.

In summary, extensive ablation studies validate the effectiveness of the proposed approach. 
Compared with direct alignment in the CLIP prior space, the constructed joint semantic space significantly improves both overall performance and semantic consistency between modality embeddings. 
Furthermore, the proposed VE-SDN framework and NGIC objective further enhance cross-modal semantic alignment during contrastive learning. 
The results demonstrate the effectiveness of VE-SDN and highlight the benefit of NGIC, particularly under the intra-subject setting.

\begin{table*}[!ht]\Huge
\renewcommand{\arraystretch}{1.2} 
\centering
\caption{Overall top-1 and top-5 accuracies (\%) of different baselines for 200-way zero-shot classification on the THINGS-EEG2 dataset.}
\label{tab:compare_baselines}

\resizebox{\linewidth}{!}{
\begin{tabular}{@{} l *{22}{c} @{}} 
\toprule
& \multicolumn{2}{c}{Subject 1} & \multicolumn{2}{c}{Subject 2} & \multicolumn{2}{c}{Subject 3} & \multicolumn{2}{c}{Subject 4} & \multicolumn{2}{c}{Subject 5} & \multicolumn{2}{c}{Subject 6} & \multicolumn{2}{c}{Subject 7} & \multicolumn{2}{c}{Subject 8}  & \multicolumn{2}{c}{Subject 9}  & \multicolumn{2}{c}{Subject 10} & \multicolumn{2}{c}{Avg} \\
\cmidrule(lr){2-3} \cmidrule(lr){4-5} \cmidrule(lr){6-7} \cmidrule(lr){8-9} \cmidrule(lr){10-11} \cmidrule(lr){12-13} \cmidrule(lr){14-15} \cmidrule(lr){16-17} \cmidrule(lr){18-19} \cmidrule(lr){20-21} \cmidrule(lr){22-23}     
Method  & top-1 & top-5 & top-1 & top-5 & top-1 & top-5 & top-1 & top-5 & top-1 & top-5  & top-1 & top-5  & top-1 & top-5  & top-1 & top-5 & top-1 & top-5  & top-1 & top-5  & top-1 & top-5  \\

\midrule 
\multicolumn{23}{c}{\textbf{Intra-subject:} train and test on one subject} \\
\midrule 

BraVL \cite{du2023decoding}  &  6.1 & 17.9  & 4.9  & 14.9 & 5.6 & 17.4 & 5.0 & 15.1  & 4.0 & 13.4 & 6.0 & 18.2 & 6.5 & 20.4 & 8.8 & 23.7 & 4.3 & 14.0 & 7.0 & 19.7 & 5.8 & 17.5  \\
NICE \cite{song2023decoding} &  12.3 & 36.6 & 10.4 & 33.9 & 13.1 & 39.0 & 16.4 & 47.0 & 8.0 & 26.9 & 14.1 & 40.6 & 15.2 & 42.1 & 20.0 & 49.9 & 13.3 & 37.1 & 14.9 & 41.9 & 13.8 & 39.5  \\
NICE-SA \cite{song2023decoding} & 13.3 & 40.2 & 12.1 & 36.1 & 15.3 & 39.6 & 15.9 & 49.0 & 9.8 & 34.4 & 14.2 & 42.4 & 17.9 & 43.6 &  18.2 & 50.2 & 14.4 & 38.7 & 16.0 & 42.8 & 14.7 & 41.7  \\
NICE-GA \cite{song2023decoding} & 15.2 & 40.1 & 13.9 & 40.1 & 14.7 & 42.7 & 17.6 & 48.9 & 9.0 & 29.7 & 16.4 & 44.4 & 14.9 & 43.1 & 20.3 & 52.1 & 14.1 & 39.7 & 19.6 & 46.7 & 15.6 & 42.8 \\
ATM-S \cite{li2024visual}  & 25.6 & 60.4 & 22.0 & 54.5 & 25.0 & 62.4 & 31.4 & 60.9 & 12.9 & 43.0 & 21.3 & 51.1 & 30.5 & 61.5 & 38.8 & 72.0 & 34.4 & 51.5 & 29.1 & 63.5 & 28.5 & 60.4   \\
\rowcolor{green!10}
\textbf{VE-SDN (Ours)}  & 34.0 & 66.7 & 34.9 & 64.6 & 41.7 & 74.3 & 43.3 & 74.8 & 29.4 & 59.5 & 42.6 & 74.1 & 34.9 & 67.7 & 52.2 & 81.5 & 38.1 & 70.7 & 45.3 & 78.4 & \textbf{39.6} & \textbf{71.2}    \\

\midrule 
\multicolumn{23}{c}{\textbf{Inter-subject:} leave one subject out for test} \\
\midrule 

BraVL \cite{du2023decoding}  &  2.3 & 8.0 & 1.5 & 6.3 & 1.4 & 5.9 & 1.7 & 6.7 & 1.5 & 5.6 & 1.8 & 7.2 & 2.1 & 8.1 & 2.2 & 7.6 & 1.6 & 6.4 & 2.3 & 8.5 & 1.8 & 7.0  \\
NICE \cite{song2023decoding} & 7.6 & 22.8 & 5.9 & 20.5 & 6.0 & 22.3 & 6.3 & 20.7 & 4.4 & 18.3 & 5.6 & 22.2 & 5.6 & 19.7 & 6.3 & 22.0 & 5.7 & 17.6 & 8.4 & 28.3 & 6.2 & 21.4 \\
NICE-SA \cite{song2023decoding} & 7.0 & 22.6 & 6.6 & 23.2 & 7.5 & 23.7 & 5.4 & 21.4 & 6.4 & 22.2 & 7.5 & 22.5 & 3.8 & 19.1 & 8.5 & 24.4 & 7.4 & 22.3 & 9.8 & 29.6 & 7.0 & 23.1  \\
NICE-GA \cite{song2023decoding} & 5.9 & 21.4 & 6.4 & 22.7 & 5.5 & 20.1 & 6.1 & 21.0 & 4.7 & 19.5 & 6.2 & 22.5 & 5.9 & 19.1 & 7.3 & 25.3 & 4.8 & 18.3 & 6.2 & 26.3 & 5.9 & 21.6  \\
ATM-S \cite{li2024visual} & 10.5 & 26.8 & 7.1 & 24.8 & \cellcolor{green!10}11.9 & \cellcolor{green!10}33.8 & \cellcolor{green!10} 14.7 & 39.4 & 7.0 & 23.9 & 11.1 & \cellcolor{green!10} 35.8 & \cellcolor{green!10} 16.1 & \cellcolor{green!10} 43.5 & \cellcolor{green!10} 15.0 & \cellcolor{green!10} 40.3 & 4.9 & 22.7 & \cellcolor{green!10} 20.5 & \cellcolor{green!10} 46.5 & 11.8 & 33.7 \\
\rowcolor{green!10}
\textbf{VE-SDN (Ours)} & 12.4 & 33.3 & 17.9 & 42.3 & \cellcolor{white}8.0 & \cellcolor{white}31.6 & \cellcolor{white}14.6 & 41.9 & 12.3 & 36.1 & 13.4 & \cellcolor{white}33.5 & \cellcolor{white}9.8 & \cellcolor{white}29.9 & \cellcolor{white}13.1 & \cellcolor{white}38.5 & 15.4 & 42.4 & \cellcolor{white}20.0 & \cellcolor{white}45.9 & \textbf{13.7} & \textbf{37.5} \\

\bottomrule

\end{tabular}}
\end{table*}

\begin{figure}[!ht]
    \centering
    \includegraphics[width=\linewidth]{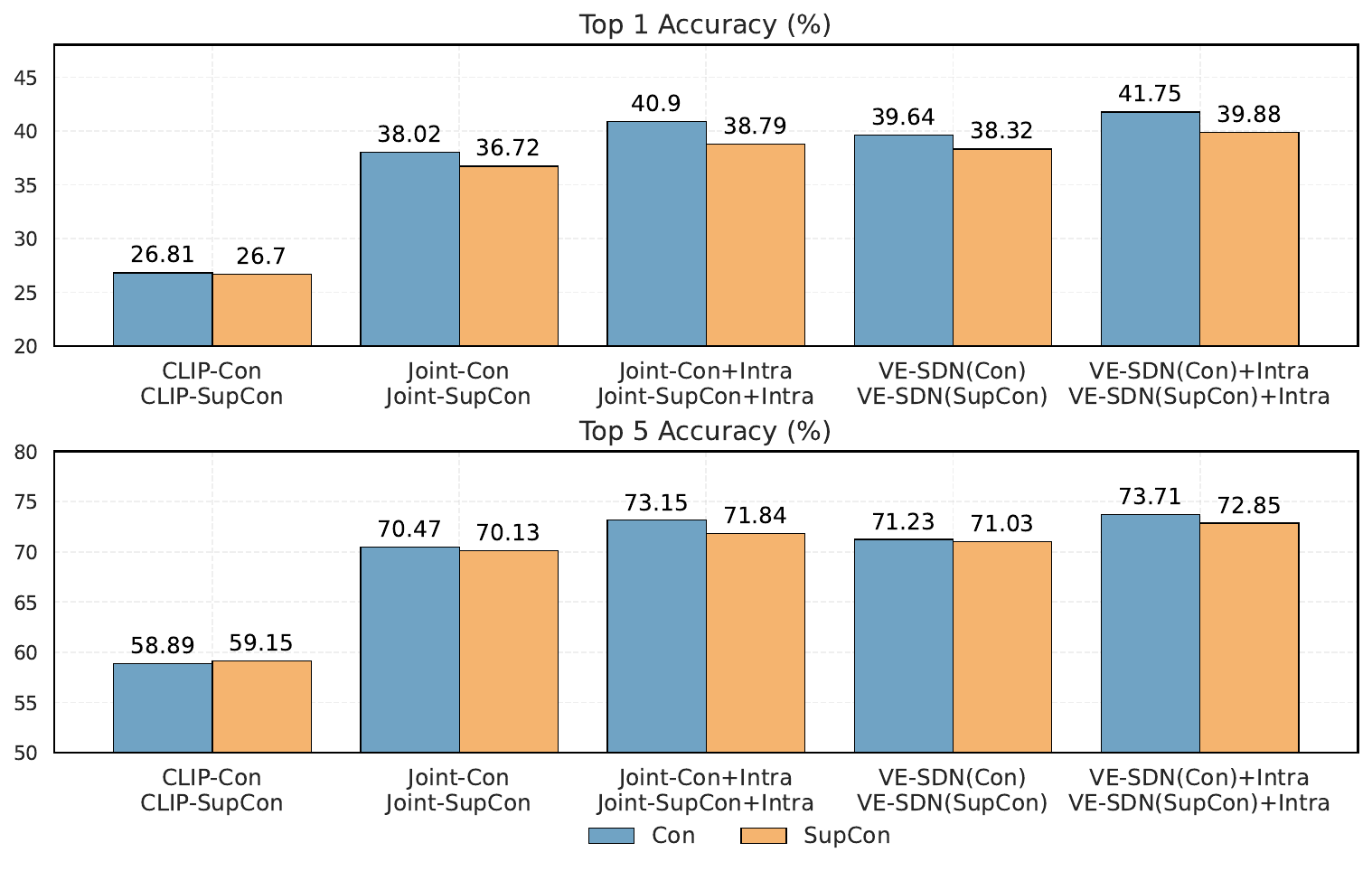}
    \caption{Comparison of average top-1 and top-5 test accuracy of Contrastive Learning and Supervised Contrastive Learning.}
    \label{fig:supcon_acc}
\end{figure}

\subsection{Comparison with Baseline Methods}
\label{sec:compare_baselines}
We compare the proposed VE-SDN with recent EEG-based neural decoding methods. For fair comparison with prior MCL-based approaches, no label information is used. The baselines are as follows:
1) BraVL \cite{du2023decoding} proposes a trimodal brain–visual–language generative framework that jointly models EEG, image, and text features via a shared latent representation.
2) NICE \cite{song2023decoding} employs a self-supervised contrastive learning framework with self-attention and graph attention to align EEG and image representations for zero-shot object recognition.
3) ATM-S \cite{li2024visual} propose the Adaptive Thinking Mapper (ATM), an EEG encoder composed of an iTransformer and spatiotemporal convolutional layers.

All classification results are presented in Table \ref{tab:compare_baselines}. 
The proposed VE-SDN consistently outperforms the strongest baseline in both intra- and inter-subject evaluations. 
Specifically, compared with ATM-S, VE-SDN achieves absolute top-1/top-5 gains of 11.1/10.8 and 1.9/3.8, corresponding to relative improvements of 38.9\%/17.9\% and 16.1\%/11.3\%, respectively.
Compared with MCL-based approaches such as NICE and ATM-S, which directly align representations in the CLIP image or text space, the inherent modality gap results in feature discrepancies and suboptimal alignment, limiting performance gains (as discussed in Section \ref{sec:ablation_study}). 
By contrast, VE-SDN aligns representations in a learned joint semantic space, where explicit semantic consistency alleviates cross-modal discrepancies and leads to further improvements in classification performance.

\subsection{Experiment on Supervised Contrastive Learning}
\label{sec:exp_on_supcon}
We exploit the label information available in Image–EEG pairs by replacing Self-Supervised Contrastive Learning with Supervised Contrastive Learning (SupCon) \cite{khosla2020supervised} to better align image and EEG representations.
Specifically, SupCon pulls together samples from different modalities that share the same class label within a mini-batch, encouraging the formation of more compact intra-class clusters in the joint embedding space.
The intra-subject experimental results using SupCon are illustrated in Fig. \ref{fig:supcon_acc}.
The results indicate that incorporating SupCon degrades the model's classification performance, rendering it overall inferior to pure contrastive learning. 
This can likely be attributed to the high intra-class variability inherent in EEG signals. 
Even for the same subject, different visual stimuli within the same semantic category can elicit significantly distinct neural responses. 
Consequently, with scarce intra-class samples and limited sample quality, forcefully compressing the cross-modal intra-class distance between images and EEG signals does not effectively enhance the robustness of representation learning.
Notably, our VE-SDN framework and NGIC approach remain effective under SupCon constraints, driving improvements in supervised classification accuracy. 
This solidly verifies the efficacy and adaptability of our proposed design.

\begin{figure*}[ht]
    \centering
    \includegraphics[width=1.0\linewidth]{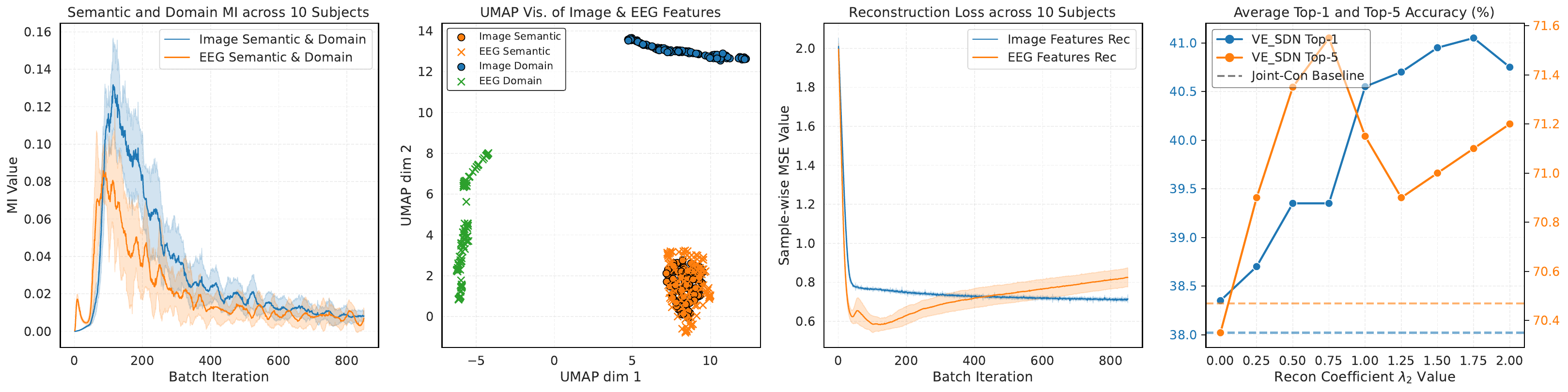} 
    \makebox[0.24\linewidth][c]{\hspace{0.0cm}(a)\hspace{-0.0cm}}%
    \makebox[0.24\linewidth][c]{\hspace{0.1cm}(b)\hspace{-0.1cm}}%
    \makebox[0.24\linewidth][c]{\hspace{0.15cm}(c)\hspace{-0.15cm}}
    \makebox[0.24\linewidth][c]{\hspace{0.3cm}(d)\hspace{-0.3cm}}
    \caption{Illustration of semantic information decoupling.
(a) Estimated MI curves between semantic and domain features during training;
(b) UMAP visualization of semantic and domain representations.
(c) Sample-wise MSE values of the cyclic reconstruction for image and EEG branches.
(d) Classification accuracy under different coefficient settings of $\lambda_2$.}
    \label{fig:mi_recloss_curve}
\end{figure*}

\begin{figure}[ht]
\centering
\includegraphics[width=\linewidth]{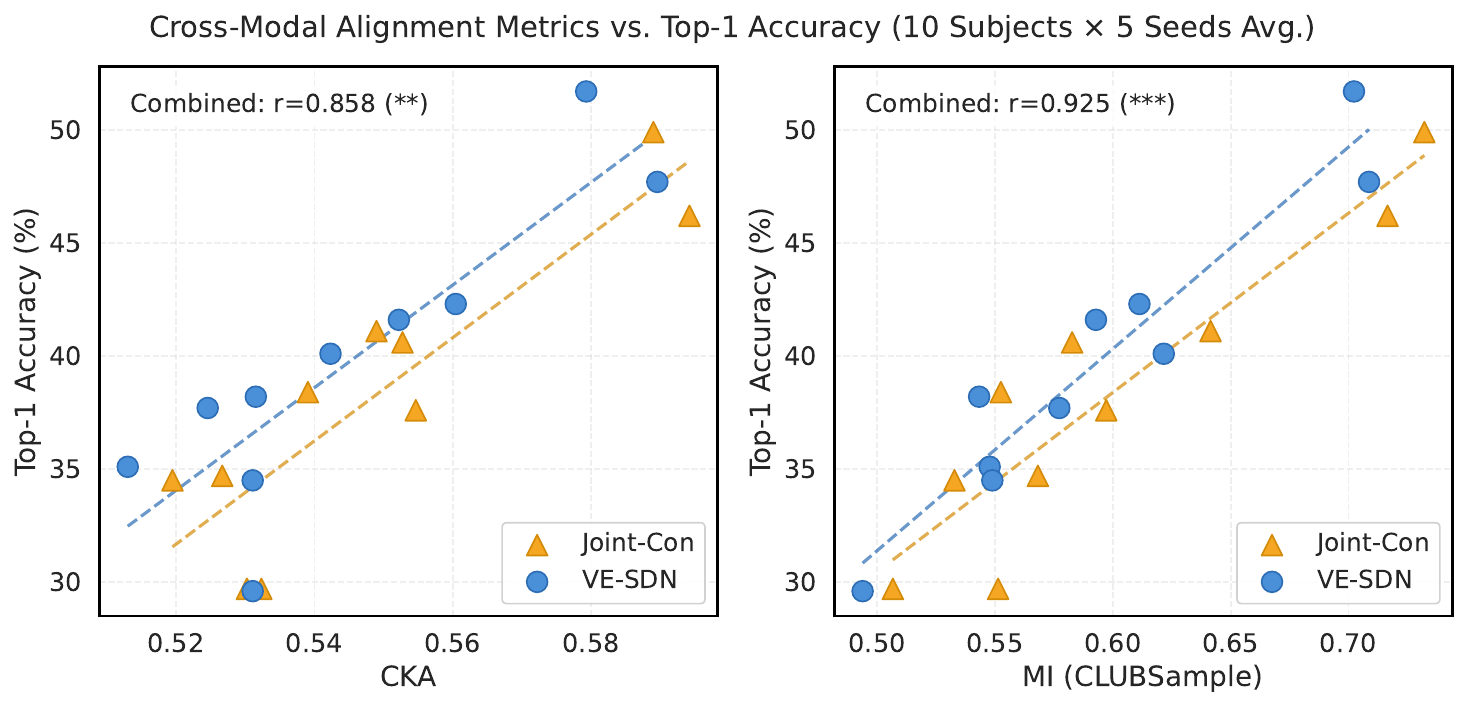}
\caption{Correlation scatter plots with linear regression lines illustrating the relationship between alignment metrics (CKA and MI) and Top-1 accuracy.}
\label{fig:pcc_ana_CKA_MI_Top1}
\end{figure}

\section{Analysis}
\label{sec:analysis}

\subsection{Analysis on Semantic Information Decoupling}
\label{sec:decouple_analysis}
In this section, we evaluate the proposed Cross-modal Semantic Information Decoupling Module.
Firstly, we plot the estimated MI curve between the modal semantic and domain features during training, along with the sample-level loss curve for cyclic reconstruction. 
As shown in Fig. \ref{fig:mi_recloss_curve} (a), the MI value between the modal semantic and domain features gradually decreases from initial fluctuations as training progresses, indicating a diminishing correlation between the features.
Fig. \ref{fig:mi_recloss_curve} (b) shows that the semantic features of images and EEG exhibit high distributional consistency, while their domain features are clearly separated, indicating successful disentanglement between semantic and domain representations.
Then, Fig. \ref{fig:mi_recloss_curve} (c) shows that the cyclic reconstruction loss of the image branch continuously decreases during training.
However, due to the information gap between the fused EEG semantic and image domain features and the original visual features, the reconstruction loss ultimately converges to a non-zero lower bound.
The reconstruction loss of the EEG branch initially decreases and then increases, due to the dynamic changes in its reconstruction target (the intermediate features output by the backbone) during training. 
In the early training stages, the intermediate features change slowly, allowing the network to fit easily, but as training progresses, the rate of change in the target exceeds the decoder's adaptation capacity, leading to a resurgence in the error.
Fig. \ref{fig:mi_recloss_curve} (d) shows that, as the weight of the cycle-consistency reconstruction increases, EEG decoding accuracy consistently improves, indicating that enhancing visual–EEG cross-modal semantic consistency effectively benefits decoding performance. When the reconstruction module is removed ($\lambda_2 = 0$), the performance drops to a level comparable to the pure contrastive learning baseline.

To further validate that the explicitly decoupled VE-SDN enhances cross-modal representation alignment and that such alignment benefits EEG decoding, we quantify the alignment between image and EEG representations on the test set and compute the Pearson correlation between alignment metrics and decoding accuracy.
We measure cross-modal alignment using Centered Kernel Alignment (CKA) \cite{kornblith2019similarity} and mutual information estimated via CLUB. 
To ensure reliable MI estimation, the CLUB network is evaluated using a 3-repeated 5-fold cross-validation protocol, where four folds are used to train the estimator and one fold is used for MI evaluation. 
The final MI score is obtained by averaging the evaluation results.

As shown in Fig. \ref{fig:pcc_ana_CKA_MI_Top1}, we plot correlation scatter plots with linear regression lines to visually illustrate the relationship between the feature alignment metrics and Top-1 accuracy across 10 subjects (averaged over 5 random seeds).
First, by jointly considering the data points from both Joint-Con and VE-SDN, we observe a significant positive Pearson correlation between the representation alignment metrics and the Top-1 accuracy. 
Specifically, the correlation coefficient between CKA and accuracy is $r = 0.858$ ($p < 0.01$), while the correlation between MI and accuracy reaches $r = 0.925$ ($p < 0.001$).
This indicates that improving cross-modal alignment between image and EEG representations can enhance EEG decoding accuracy.
Additionally, as shown in Fig. \ref{fig:pcc_ana_CKA_MI_Top1}, the linear regression fit for the VE-SDN data points is situated noticeably higher than that of Joint-Con. 
This demonstrates that VE-SDN can further enhance cross-modal representation alignment, which directly translates into superior EEG decoding performance.
Notably, consistent with recent findings \cite{almudevar2025aligning} that filtering modality-specific information enhances alignment, VE-SDN leverages explicit decoupling to strip away modality noise and purify shared semantics, ultimately achieving superior representation alignment and decoding performance.




\begin{table}[!t]
\centering
\small
\caption{Average accuracy (\%) of NGIC objective with different distance metrics.}
\label{tab:ab_dist}
\setlength{\tabcolsep}{6pt}
\renewcommand{\arraystretch}{1.15} 
\begin{tabular}{lcccc}
\toprule
& 
\multicolumn{2}{c}{Joint-Con} &
\multicolumn{2}{c}{VE-SDN} \\
\cmidrule(lr){2-3} \cmidrule(lr){4-5}

\rlap{Distance}\phantom{NGIC (EProto-V2E)} & Top-1 & Top-5 & Top-1 & Top-5 \\
\midrule
 $\ell_1$ Distance  & 35.73 & 67.41 & 34.94 & 66.91 \\
Euclidean         & 40.51 & 73.08 & 41.50 & 72.15 \\
Cosine            & \textbf{40.90} & \textbf{73.15} & \textbf{41.75} & \textbf{73.71} \\
\bottomrule
\end{tabular}
\end{table}

\begin{table}[!t]
\centering
\small
\caption{Average accuracy (\%) of Intra-class consistency based on different prototypes.}
\label{tab:ab_proto}
\setlength{\tabcolsep}{6pt}
\renewcommand{\arraystretch}{1.15} 
\begin{tabular}{lcccc}
\toprule
 &
\multicolumn{2}{c}{Joint-Con} &
\multicolumn{2}{c}{VE-SDN} \\
\cmidrule(lr){2-3} \cmidrule(lr){4-5}
Method & Top-1 & Top-5 & Top-1 & Top-5 \\
\midrule
2Proto-Cen   & 38.49 & 71.42 & 39.52 & 71.58 \\
Share-Proto      & 39.20 & 72.53 & 40.04 & 72.41 \\
VProto-E2V   & 38.23 & 70.89 & 37.99 & 70.70 \\
NGIC (EProto-V2E)   & \textbf{40.90} & \textbf{73.15} & \textbf{41.75} & \textbf{73.71} \\
\bottomrule
\end{tabular}
\end{table}

\begin{figure}[!t]
\centering
\includegraphics[width=\linewidth]{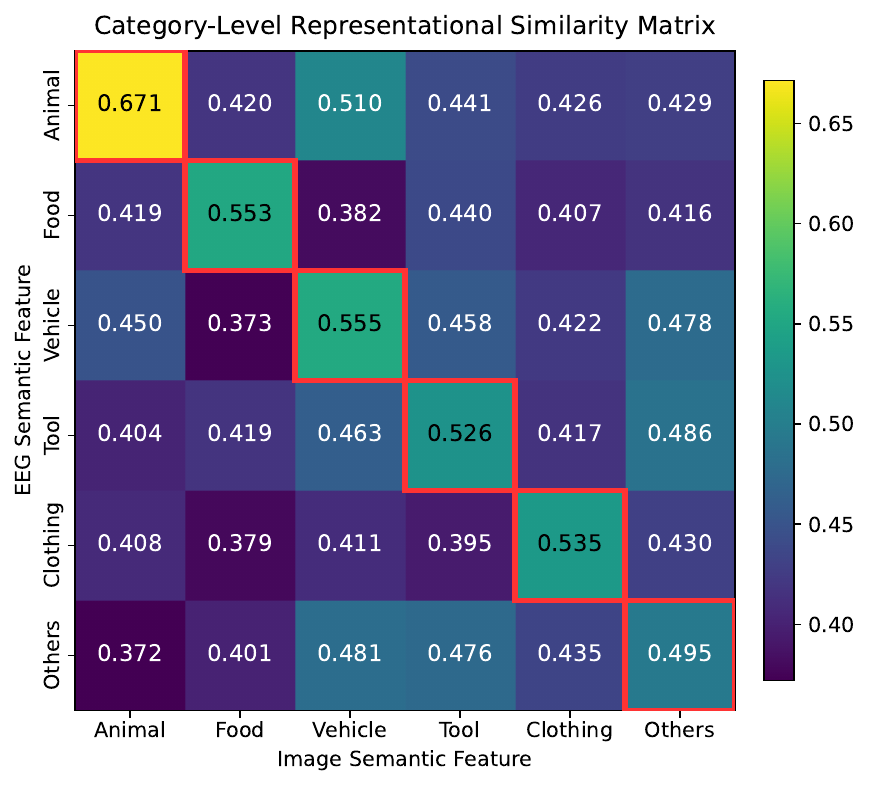} 
\caption{Representation similarity matrices for EEG and image semantic features across six subcategories within 200 test concepts.}
\label{fig:rsa_heat_map}
\end{figure}

\begin{figure}[!t]
\centering
\includegraphics[width=\linewidth]{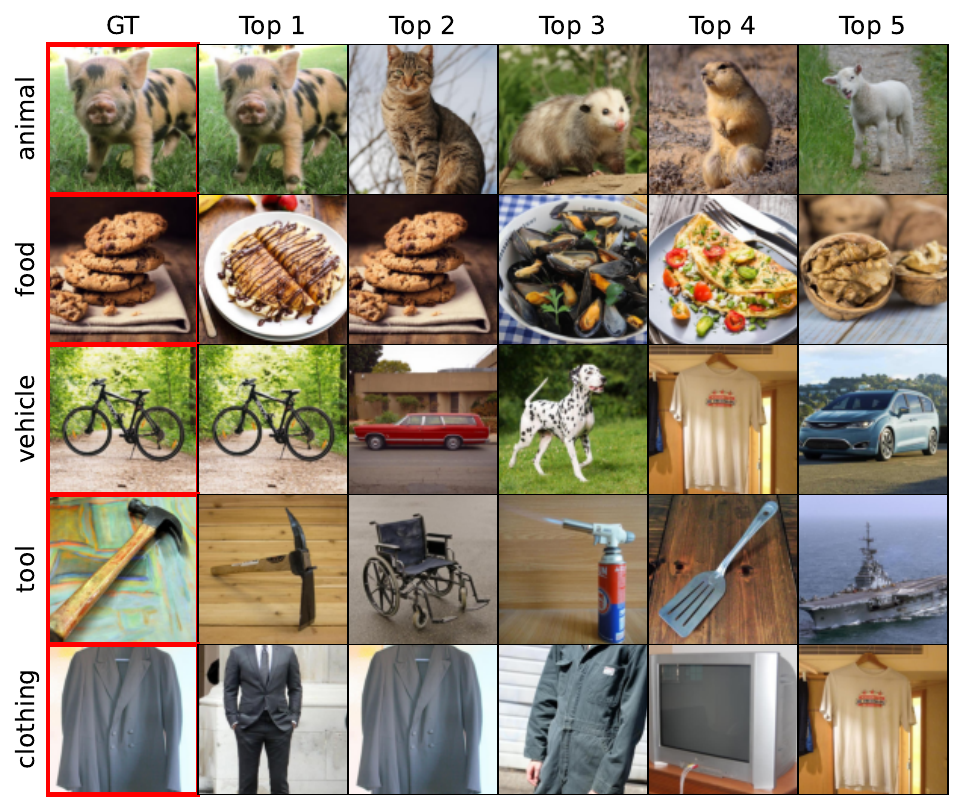}
\caption{Top-5 image retrieval results for five subcategories in the test set.}
\label{fig:retrieval_images}
\end{figure}

\subsection{Analysis of Intra-Class Consistency}
\label{sec:intra_class}
We aim to leverage label information to further enhance representation learning and thereby improve the decoding performance of EEG signals. 
However, as demonstrated in Section \ref{sec:exp_on_supcon}, directly applying supervised contrastive learning to enforce tighter alignment between EEG signals and stimulus images from the same category does not lead to improved decoding performance.
A possible explanation is the high intra-class variability of EEG signals.
Therefore, to obtain robust class-level EEG representation, we maintain an EEG class prototype during training and reduce the distance between image embeddings of the same category and their corresponding EEG prototypes. 
The performance under different distance metrics is reported in Table \ref{tab:ab_dist}.
The results show that cosine distance, which aligns with the optimization objective of the main contrastive loss, achieves the best performance.
Additionally, we further compare several methods related to intra-class consistency. 
Inspired by Center Loss \cite{wen2016discriminative}, we design Two-Prototype Centering (2Proto-Cen), where each modality maintains its own class prototype and pulls its embeddings toward the corresponding center. 
Based on Cross-modal Center Loss \cite{jing2021cross}, we introduce Shared-Prototype (Share-Proto), where EEG and image embeddings share a common class prototype and are jointly aligned to it.
We also propose Visual-Prototype Anchoring (VProto-E2V), which pulls EEG embeddings toward visual class prototypes. 
Finally, our method, NGIC (EProto-V2E), maintains EEG class prototypes and aligns image embeddings toward them.
As shown in Table~\ref{tab:ab_proto}, our NGIC strategy achieves the best performance among all compared methods.
For the variants of intra-class consistency methods described above, we observe the following: 2Proto-Cen only enhances intra-modality consistency without altering cross-modal relationships, resulting in limited performance improvement.
Share-Proto improves cross-modal intra-class consistency but also modifies the representation structures of both EEG and Image, potentially interfering with the alignment learned by the contrastive loss.
The VProto-E2V method fails to form an effective EEG class prototype representation and does not establish stable cross-modal matching relationships, potentially impairing EEG decoding performance, as in simple supervised contrastive learning.
In contrast, NGIC (EProto-V2E) applies constraints solely on the Image embeddings, leaving the EEG representations fully governed by the contrastive loss.
This asymmetric constraint effectively enhances the stability of cross-modal matching and EEG-based decoding performance.


\subsection{Semantic Similarity and Image Retrieval}
To verify that our proposed VE-SDN significantly enhances the semantic consistency between cross-modal representations, we conducted two visualization analyses: (1) semantic representation similarity analysis, and (2) image retrieval visualization.
We subdivide the 200 test object concepts into six major categories: animal, food, vehicle, tool, clothing, and others. 
We utilize subject 8 for analysis and calculate the average semantic feature similarity matrix across six categories.
The resulting category-level representational similarity matrix is shown in Fig. \ref{fig:rsa_heat_map}.
As observed, the average similarity within the same semantic category is consistently higher than that between different categories, indicating that the proposed VE-SDN effectively captures and aligns semantic category features. 
Next, we conducted an EEG-based image retrieval analysis. 
Specifically, we randomly sampled EEG features from five subcategories to serve as queries and presented the Top-5 matched images. 
As shown in Fig. \ref{fig:retrieval_images}, the retrieved results exhibit highly semantic correlation with the ground truth, further validating the effectiveness of the VE-SDN in cross-modal semantic feature retrieval.

\section{Discussion, Limitations, and Conclusion}
\label{sec:dis_lim_con}
\subsection{Discussion}
In this study, we construct a Visual–EEG joint semantic space, in which frozen visual representations and extracted EEG representations are projected through nonlinear mappings and aligned via MCL. 
Compared with previous methods that directly align features in the pretrained space, the proposed Joint-Con strategy not only improves decoding performance significantly, but also alleviates the inherent modality gap in cross-modal alignment. 
This finding underscores the importance of a well-designed latent space for bridging modality discrepancy and facilitating effective EEG neural decoding.
Building upon this, we propose the Visual–EEG Semantic Decoupling Network to strengthen semantic consistency between image and EEG representations during alignment.
Quantitative metrics on test samples, including CKA and MI, show that VE-SDN improves cross-modal alignment, which in turn enhances decoding performance. 
This suggests that explicitly improving representation alignment, such as through mutual information maximization, can benefit neural decoding.
Furthermore, we designed a Neural-Guided Intra-Class Consistency objective. 
By fully leveraging label information, it introduces a tailored asymmetric training guidance to further enhance the representation consistency from the visual to the EEG modality.

These results indicate that explicitly modeling semantic consistency between visual and neural representations in cross-modal learning can significantly enhance EEG-based visual decoding, potentially benefiting practical BCI systems and neural representation analysis.

\subsection{Limitations and Future Work}
Although the proposed approach demonstrates strong decoding performance, it still has several limitations, as follows:
\begin{enumerate}
\item \textbf{Limited inter-subject generalization:} 
Inter-subject experimental results indicate that although VE-SDN extracts more generalizable representations and achieves optimal decoding performance, the overall improvement is still limited.
As observed in the inter-subject evaluation of VE-SDN+NGIC, extracted EEG semantic features still retain substantial subject-dependent information.
This indicates that without explicit cross-subject alignment constraints, completely disentangling subject-invariant semantic features from highly individualized EEG signals remains a severe challenge.

\item \textbf{Limited focus on low-level visual details:} 
Our approach focuses on aligning high-level cross-modal semantics. 
While this strategy benefits EEG-based classification and retrieval, it may be suboptimal for fine-grained downstream tasks, such as pixel-level visual reconstruction or multi-granularity neural signal analysis.

\item \textbf{Unverified across multiple datasets:} Our evaluation is confined to a single visual-evoked EEG dataset. 
The adaptability of our approaches to diverse EEG datasets or paradigms remains untested.
\end{enumerate}

To overcome these limitations and advance visual neural decoding, our future work will focus on:
\begin{enumerate}
\item \textbf{Subject-invariant feature disentanglement:} Future work will focus on disentangling subject-invariant features to overcome the bottleneck of cross-subject generalization, which is critical for practical BCIs deployment.

\item \textbf{High-fidelity visual reconstruction:} We plan to extend our approaches to capture both high-level semantics and low-level perceptual details, thereby enabling high-fidelity, pixel-level EEG-based visual reconstruction.

\item \textbf{Extensive cross-dataset validation:} 
We will rigorously validate our framework across a broader range of EEG datasets \cite{xu2024alljoined1} and experimental paradigms to ensure its generalizability and broader applicability.
\end{enumerate}

\subsection{Conclusion}
This study effectively narrows the inherent modality gap between visual and EEG representations in cross-modal alignment, and alleviates the sub-optimal cross-modal alignment.
Specifically, we construct a Visual-EEG joint semantic space to bridge the gap between visual images and neural signals. 
Building upon this foundation, we propose two novel approaches: (1) a Visual-EEG Semantic Decoupling Network and (2) a Neural-Guided Intra-Class Consistency objective, to explicitly enhance semantic consistency during the representation alignment process.
Extensive ablation and comparative experiments demonstrate the effectiveness of the proposed modules and show that they consistently outperform baselines.
This performance enhancement in visual-evoked EEG decoding also holds immense potential for the development of emerging BCIs.

\bibliographystyle{IEEEtran}
\bibliography{IEEEabrv, myreferences}

\end{document}